\documentclass[11pt,a4paper]{article}
\usepackage{authblk}

\usepackage[hyperref]{ranlp2025}
\usepackage{times}
\usepackage{latexsym}
\usepackage{adjustbox}
\usepackage{booktabs}
\usepackage{caption}
\usepackage{hyperref}
\usepackage{tcolorbox}
\usepackage{graphicx}
\usepackage{multirow}
\usepackage{makecell}
\usepackage{tabularx}
\usepackage{xcolor}
\usepackage{listings}
\usepackage{amsmath}
\usepackage{tikz}
\usepackage{pgfplots}
\usepackage{pgf-pie}
\usepackage{subcaption}
\usepackage{enumitem}
\usetikzlibrary{positioning, shapes, arrows.meta}

\definecolor{plotblue}{HTML}{1f77b4}
\definecolor{plotorange}{HTML}{ff7f0e}
\definecolor{plotgreen}{HTML}{2ca02c}
\definecolor{plotred}{HTML}{d62728}
\definecolor{plotpurple}{HTML}{9467bd}

\definecolor{myhl}{RGB}{255,235,59}

\usepackage{microtype}

\aclfinalcopy 

\title{Evaluating Structured Decoding for Text-to-Table Generation: Evidence from Three Datasets}

\author[1]{Julian Oestreich}
\author[1,2]{Lydia Müller}
\affil[1]{Institute for Applied Informatics (InfAI) at Leipzig University}
\affil[2]{Leipzig University}

\date{15. July 2025}

\begin{document}
\maketitle
\begin{abstract}
We present a comprehensive evaluation of structured decoding for text-to-table generation with large language models (LLMs). While previous work has primarily focused on unconstrained generation of tables, the impact of enforcing structural constraints during generation remains underexplored. We systematically compare schema-guided (structured) decoding to standard one-shot prompting across three diverse benchmarks - E2E, Rotowire, and Livesum - using open-source LLMs of up to 32B parameters, assessing the performance of table generation approaches in resource-constrained settings. Our experiments cover a wide range of evaluation metrics at cell, row, and table levels. Results demonstrate that structured decoding significantly enhances the validity and alignment of generated tables, particularly in scenarios demanding precise numerical alignment (Rotowire), but may degrade performance in contexts involving densely packed textual information (E2E) or extensive aggregation over lengthy texts (Livesum). We further analyze the suitability of different evaluation metrics and discuss the influence of model size.
\end{abstract}

\section{Introduction}

Automatically converting text into structured tables has become a key challenge in information extraction and data-driven reporting. By converting unstructured content into tables, downstream tasks such as knowledge‐base construction \cite{table-to-kg, tab2know}, document summarization, and web chatbot readability \cite{mdeval} can be improved. Early work framed the task as a sequence-to-sequence learning problem using encoder-decoder architectures~\cite{wu-etal-2022-text-table}, while the more recent approaches leverage LLMs with different prompting techniques, often over multiple stages. While recent advances in constrained decoding and grammar-based generation have led to improvements in structured output tasks, these methods have not yet been systematically applied to the text-to-table task. As a result, the impact of enforcing structural constraints during generation remains underexplored. Our contribution is as follows: We compare schema-guided decoding to one-shot prompting on E2E, Rotowire, and Livesum to assess how schema enforcement impacts validity and semantic quality of the resulting markdown tables at the cell, row, and table levels. We also examine model-size effects using open-source LLMs up to 32B parameters and evaluate metric suitability to guide future text-to-table research. 


\section{Related work}

The Text-to-Table generation task was introduced by \cite{wu-etal-2022-text-table}, who framed it as a sequence-to-sequence learning problem within the field of information extraction. They were using fine-tuned BART-based models on pairs of texts and tables to predict table representations based on the textual input. Further works have been using Large Language Models (LLMs) to solve the problem, with key differences regarding the type of generated markup sequence for table representation \cite{tang-etal-2024-struc}, their prompting techniques \cite{coyne2024llm2table}, underlying datasets and whether the overall table format (schema) was provided to the model or not. The majority of the works was providing the table schema either while training or prompting \cite{coyne2024llm2table, jiao-etal-2023-instruct, tang-etal-2024-struc}, while the recent work of \citet{ahuja2025mapmakeschemaguidedtext} did not. While these prior studies have advanced text-to-table generation using either large-scale proprietary models or fine-tuned open-source LLMs—often ranging from 7B to 70B parameters or more, our work systematically investigates schema-guided decoding in smaller, publicly available open-source models. Recent advances in generative artificial intelligence have shown, that the generation of structured outputs, such as tables, could benefit from constrained decoding strategies \cite{grammar-aligned-decoding, geng-etal-2023-grammar}. Notably, \citet{geng-etal-2023-grammar} found that "grammar-constrained LMs substantially outperform unconstrained LMs" on structured NLP tasks like information extraction and constituency parsing, while \citet{tam-etal-2024-speak} found, that these decoding strategies degrade the semantical correctness of generated outputs. Most existing evaluations focus on general text generation or specialized NLP tasks, suggesting that the impact of constrained decoding is task-dependent. This leaves gaps in our understanding of how constraints affect content generation for structured data representations, such as tabular data.

\section{Methodology}

\subsection{Data}
\setlength{\aboverulesep}{0pt}
\setlength{\belowrulesep}{0pt}
\begin{table*}[]
  \centering
  \renewcommand{\arraystretch}{1.2}
  \small
  \resizebox{\textwidth}{!}{%
    \begin{tabular}{|l | r | ccc | ccc | r | ccc |}
      \toprule
      \multicolumn{1}{|c|}{} 
        & \multicolumn{7}{c|}{\textbf{Table}} 
        & \multicolumn{4}{c|}{\textbf{Input Text}} \\
      \cmidrule{2-8} \cmidrule{9-12}
      Dataset 
        & N & Rows min & Rows max & Rows mean & Cols min & Cols max & Cols mean
        & N & Word min & Word max & Word mean \\
      \midrule
      E2E         & 4693 & 2  & 2  & 2.00 & 1 & 7  & 5.40 & 4693 & 4   & 71   & 24.06  \\
      RW Teams    & 687  & 2  & 3  & 2.91 & 1 & 9  & 4.22 
        & \multirow{2}{*}{728} & \multirow{2}{*}{135} & \multirow{2}{*}{695} & \multirow{2}{*}{308.36} \\
      RW Players  & 724  & 2  & 16 & 7.49 & 1 & 17 & 7.94 &      &      &      &        \\
      LiveSum     & 754  & 3  & 3  & 3.00 & 9 & 9  & 9.00 & 754  & 724 & 1760 & 1138.40 \\
      \bottomrule
    \end{tabular}%
  }
  \caption{Descriptive statistics of the gold table data and the input texts for each dataset. For the Rotowire dataset we show two rows that correspond to the different table types (Teams and Players).}
  \label{tab:datasets_descr}
\end{table*}

Our benchmarking data consists of three datasets, each with distinct characteristics spanning a spectrum of text-to-table challenges: the E2E dataset \cite{novikova-etal-2017-e2e}, the Rotowire dataset \cite{wiseman-etal-2017-challenges}, and the Livesum dataset \cite{text-tuple-table}. E2E and Rotowire were originally designed for the Table-to-Text generation task, while Livesum was specifically created for the Text-to-Table task. E2E and Rotowire, along with WikiTableText \cite{wikitabletext} and WikiBio \cite{wikibio}, were repurposed for Text2Table evaluation by \citet{wu-etal-2022-text-table}. However, aside from Rotowire, these repurposed datasets lack structural diversity, as they only feature simple tables with two columns. For this reason, we only include E2E as a representative of simple tables in our experiments. 

Descriptive statistics on the datasets show that while the input texts for E2E are very short, with an average of 24 words, the sizes in the other two datasets are significantly larger: Rotowire with an average of 308 words, and Livesum with 1,138 words on average (see Table \ref{tab:datasets_descr}). While E2E and Livesum show a uniform distribution of row- and column sizes, the Rotowire tables have a greater diversity regarding their sizes. E2E, sourced from the restaurant domain, consists of short textual descriptions paired with two-row tables summarizing restaurant attributes. Its focus lays on extracting textual information from short texts into simple tables. The Rotowire dataset originates in the sport domain and is a widely used benchmark in natural language generation and information extraction \cite{neural_methods_2024,content_selection_planning}. Each example contains one or more tables of statistics on basketball players and teams (e.g., points, assists, rebounds), paired with human-written game summaries and it requires the identification and assignment of sparsely mentioned numerical statistics to player and team tables. Livesum \cite{text-tuple-table} also comes from the sports domain and comprises live soccer commentaries together with team statistics. It demands the truthful aggregation of atomic extraction units, such as individual events (e.g., goals, fouls), that are distributed throughout a longer text and organize them into comprehensive tables, a task that likely demands enhanced reasoning capabilities of the models. 

\begin{figure}[h!]
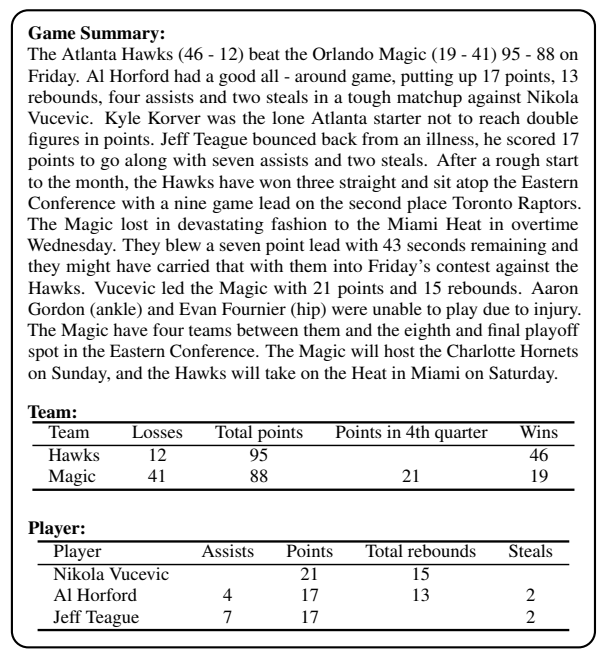

    \centering
        \begin{tcolorbox}[colback=white, colframe=black, boxrule=0.8pt, arc=2mm, left=1mm, right=1mm, top=1mm, bottom=1mm]
    \begin{scriptsize}
        \textbf{Game Summary:} \\
        The Atlanta Hawks (46 - 12) beat the Orlando Magic (19 - 41) 95 - 88 on Friday. Al Horford had a good all - around game, putting up 17 points, 13 rebounds, four assists and two steals in a tough matchup against Nikola Vucevic. Kyle Korver was the lone Atlanta starter not to reach double figures in points. Jeff Teague bounced back from an illness, he scored 17 points to go along with seven assists and two steals. After a rough start to the month, the Hawks have won three straight and sit atop the Eastern Conference with a nine game lead on the second place Toronto Raptors. The Magic lost in devastating fashion to the Miami Heat in overtime Wednesday. They blew a seven point lead with 43 seconds remaining and they might have carried that with them into Friday's contest against the Hawks. Vucevic led the Magic with 21 points and 15 rebounds. Aaron Gordon (ankle) and Evan Fournier (hip) were unable to play due to injury. The Magic have four teams between them and the eighth and final playoff spot in the Eastern Conference. The Magic will host the Charlotte Hornets on Sunday, and the Hawks will take on the Heat in Miami on Saturday.

        \vspace{1em}
        \textbf{Team:} \\
        \begin{adjustbox}{max width=\columnwidth, center}
        \begin{tabular}{lcccc}
            \toprule
            Team & Losses & Total points & Points in 4th quarter & Wins \\
            \midrule
            Hawks  & 12 & 95 &  & 46 \\
            Magic  & 41 & 88 & 21 & 19 \\
            \bottomrule
        \end{tabular}
        \end{adjustbox}
        
        \vspace{1.5em}
        
        \textbf{Player:} \\
        \begin{adjustbox}{max width=\columnwidth, center}
        \begin{tabular}{lcccc}
            \toprule
            Player & Assists & Points & Total rebounds & Steals \\
            \midrule
            Nikola Vucevic &   & 21 & 15 &  \\
            Al Horford     & 4 & 17 & 13 & 2 \\
            Jeff Teague    & 7 & 17 &  & 2 \\
            \bottomrule
        \end{tabular}
        \end{adjustbox}
    \end{scriptsize}
    \end{tcolorbox}
    \caption{Example from the Rotowire dataset showing a game summary with the corresponding team and player box scores.}
    \label{fig:rotowire-fullboxed}
\end{figure}


\subsection{Prompting \& Generation}

For generating structured tables from text using Large Language Models (LLMs), we follow two different methods: free-form (unstructured) generation and schema-guided (structured) decoding.

In the free-form approach, LLMs generate a markdown sequence for a given input text. A one-shot instruction prompt, empirically refined through iterative experimentation, encourages the model to adhere to the desired table output format. Specifically, we provide the intended table structure in the prompt, specifying header cells as well as the number of columns and rows, inspired by \citet{tang-etal-2024-struc}.

In contrast, the schema-guided approach enforces tighter structural guarantees through decoding constraints defined by a provided JSON schema. To enable constrained decoding for structured table generation, we implement a schema builder that dynamically constructs a nested JSON schema based on the table layout specified by the row and column headers in the gold data. Cell values are represented as nullable integers, while both row and column headers are constrained to predefined values.

\begin{figure}[h]
\centering
\begin{adjustbox}{width=\columnwidth}
\begin{tikzpicture}[
    box/.style={draw, rounded corners, font=\normalsize, align=center, minimum height=0.9cm, minimum width=3.7cm, inner sep=4pt},
    model/.style={draw, rectangle, font=\normalsize, align=center, minimum width=3.8cm, minimum height=1.4cm},
    decbox/.style={draw, rectangle, font=\normalsize, fill=gray!10, align=center, minimum width=3.2cm, minimum height=0.7cm, inner sep=2pt},
    tablebox/.style={ font=\normalsize, align=center, minimum width=4.2cm, minimum height=5.5cm},
    arrow/.style={-{Stealth}, thick},
    node distance=0.4cm and 1.3cm
]

\node[box] (promptA) {ONE-SHOT\\Instruction Prompt};
\node[model, below=0.9cm of promptA] (modelA) {Model};
\node[tablebox, below=2cm of modelA, text width=5cm] (tableA) {
\textbf{Team Table}\\
\begin{tabular}{|c|c|c|}
\hline
TEAM & Wins & Pts \\
\hline
Hawks & 46 & 95 \\
Magic & 19 & 88 \\
\hline
\end{tabular}\\[0.2em]
\textbf{Player Table}\\
\begin{tabular}{|c|c|c|}
\hline
PLAYER & Points & Ast. \\
\hline
A. Horford & 17 & 4 \\
J. Teague & 17 & 7 \\
N. Vucevic & 21 &  \\
\hline
\end{tabular}
};

\node[box, right=3.8cm of promptA] (promptB) {Guided Decoding Prompt};
\node[model, below=0.9cm of promptB, align=center, minimum height=2.1cm,inner sep=2pt] (modelB) {Model\\[0.6cm]};

\node[decbox, below=+1.0cm of modelB.north] (decoderB) {Constrained Decoder};

\node[tablebox, below=1.5cm of modelB, text width=5.8cm, align=left] (jsonB) {
\normalsize
\begin{tabular}{l}
\texttt{\{} \\
\ \ \texttt{"Team": \{} \\
\ \ \ \ \texttt{"Hawks": \{} \\
\ \ \ \ \ \ \texttt{"wins": 46, ...} \\
\ \ \ \ \texttt{\},} \\
\ \ \ \ \texttt{...} \\
\ \ \texttt{\},} \\
\ \ \texttt{"Player": \{} \\
\ \ \ \ \texttt{"Nikola Vucevic": \{} \\
\ \ \ \ \ \ \texttt{"points": 21,} \\
\ \ \ \ \ \ \texttt{"assists": null, ...} \\
\ \ \ \ \texttt{\},} \\
\ \ \ \ \texttt{...} \\
\ \ \texttt{\}} \\
\texttt{\}}
\end{tabular}
};
\node[below=0.13cm of jsonB] (captionB) {(b) Guided Decoding w/ Structured JSON};
\node[left=2cm of captionB] {(a) Open Prompt w/ Markdown Tables};

\node[box, minimum width=2cm, minimum height=1.5cm, text width=1.2cm, inner sep=1pt, left=0.9cm of decoderB, font=\small, align=center] (schemaB) {
\textbf{Pythonic\\JSON\\Schema}
};
\draw[arrow] (promptA) -- (modelA);
\draw[arrow] (modelA) -- (tableA);

\draw[arrow] (promptB) -- (modelB);
\draw[arrow] (modelB) -- (jsonB);
\draw[arrow] (schemaB.east) -- (decoderB.west);

\end{tikzpicture}
\end{adjustbox}
\caption{Comparison of open prompting and guided decoding. In guided decoding, the decoder component is constrained to a Pythonic JSON schema.}
\label{fig:prompt-vs-guided}
\end{figure}
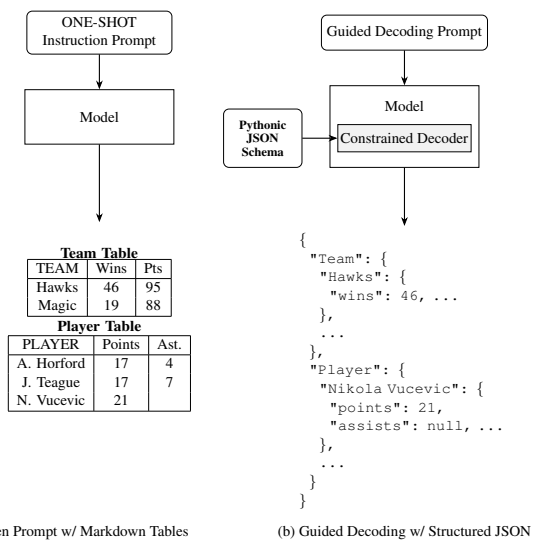

\subsection{Parsing \& Post-processing}

The postprocessing step of the schema guided approach is comparatively simple. The LLM emits JSON with Pydantic-safe keys, while the schema keeps the true column names in each property’s title metadata. We swap every key for its title (e.g., \texttt{total\_points} → `Total Points'), then turn each top-level object into a pandas DataFrame.

In case of unstructured (free-form) generation of markdown tables, Large language models (LLMs) often ignore rigid “table-only” instructions. Even when explicitly asked to emit nothing but a Markdown table, they may

\begin{itemize}
  \item prepend or append free-form prose,
  \item break a single table with stray blank lines or malformed rows, or
  \item return \emph{several} distinct tables in succession.
\end{itemize}

To robustly recover well-formed tabular data under all of these failure modes, we adopt a two-stage, candidate-based pipeline: we first \emph{extract} every pipe-delimited region that \emph{could} be a table, then \emph{validate} each region against Markdown’s structural rules. This separation lets us (i) retain all legitimately produced tables, regardless of how many the model generates. This is particularly important for the Rotowire dataset, where the expected output frequently consists of two tables; and (ii) pinpoint exactly where and why a malformed candidate breaks the specification.

\paragraph{1. Candidate extraction.}
We scan the raw LLM output line-by-line, grouping together each maximal run of lines that begin with a pipe (\texttt{|}). Every such run becomes a \emph{candidate block}: it might be a complete table, a fragment, or just arbitrary pipe-separated text. Because we postpone any judgment of correctness, no genuine table can be missed.

\paragraph{2. Table validation \& parsing.}
Each candidate block is subjected to four sequential regex checks:

\begin{enumerate}[label=(\roman*)]
  \item \textbf{Header integrity.} The first line must start and end with \texttt{|} and contain at least one non-pipe character between consecutive pipes.
  \item \textbf{Separator row.} The second line must also be pipe-delimited and include at least three hyphens per column (optionally flanked by colons), satisfying Markdown’s header-body separation rule.
  \item \textbf{Row consistency.} Every subsequent line must open and close with \texttt{|} and, when split on pipes, yield the same number of cells as the header, guaranteeing a uniform column alignment.
  \item \textbf{Table size.} Each table must have at least three rows: One header, one separator and one data row.
\end{enumerate}

Only candidates that pass all four checks are split on pipes, trimmed of whitespace, and assembled into a structured grid (e.g., a \texttt{DataFrame}). Failures, such as missing pipes, malformed separators, or mismatched cell counts, raise specific errors, enabling fine-grained diagnostics of LLM formatting bugs.

Because the extraction and validation steps operate on each candidate independently, the pipeline naturally recovers valid tables in their original order, even when a single model invocation produces multiple tables. 

\subsection{Evaluation}

For constructing a mapping between table candidates and gold table we are following a greedy table assignment approach. For each candidate table we are scoring the overlap between its column headers and the gold table using case-insensitive matching. The candidate table with the highest score is then assigned for evaluation and if there is not at least one table with at least one overlapping column header found in the candidates for evaluation, we mark the table as missing.

\subsubsection{Metrics}

For assessing the quality of the generated tables we report performance at three granularity levels: Cell-Level, Row-Level, and Table-Level. At the strictest level, we measure \emph{table accuracy}, defined as the proportion of tables where every normalized predicted cell exactly matches the corresponding gold cell. At the \emph{row-level}, we count a true positive (TP) as a predicted row that exactly matches a gold row (order-agnostic), a false positive (FP) as an extra predicted row, and a false negative (FN) as a missing gold row—suitable when rows represent unique entities (e.g., player or team statistics). At the \emph{cell-level}, TP occurs when predicted and gold values exactly match after normalization, FP when a predicted cell exists but the gold cell does not, and FN when a gold value exists but the prediction is missing or incorrect. We then report F1 scores at both cell and row levels. 

We calculate the ROUGE-L Score \cite{lin-2004-rouge} and the Levenshtein ratio, a normalized measure derived from Levenshtein distance \cite{levenshtein1966binary}, to quantify string similarity upon table level, after transforming the DataFrame into a table sequence, by deterministically applying a minimal markdown table format. We also calculate the Levenshtein ratio positionally on cell-level and calculate the average over every (non-header) cell, treating missing cells as empty strings. 

For the Rotowire and Livesum datasets, we further calculate Root Mean Square Error (RMSE), since their inner cells are numeric. All metrics exclude header cells, as these are always provided by the schema-containing prompts. The metrics are only calculated on tables that were present in the generated output. Missing tables therefore do not influence the calculation per metric, but we keep track of the actual presence of expected tables in the generated outputs.

\subsection{Experimental Setup}

Our experiments were performed on a single node of the high‑performance computing cluster at the scientific computing cluster of the University of Leipzig\footnote{\url{https://www.sc.uni-leipzig.de/}}. 
The node contains two AMD(R) EPYC(R) 7713 CPUs with 64 cores each, 1TB RAM and eight Nvidia A30 GPUs, each with 24GB HBM2 RAM. For the generation we leverage the \texttt{vLLM} library \cite{kwon2023efficient}, with a unified setup over all evaluated models: A \texttt{temperature} of 0.0, a \texttt{max\_model\_len} of 6144 and \texttt{max\_new\_tokens} of 4096. For structured decoding we are using the \texttt{xgrammar} package. The code we used for generating and evaluating our models was made available in our Gitlab Repository\footnote{\url{https://github.com/JulianOestreich90/text2table}}.

\section{Results}

\subsection{E2E}

\begin{table*}[h!]
  \centering
  \small
  \renewcommand{\arraystretch}{1.2}
  \resizebox{\textwidth}{!}{%
    \begin{tabular}{|l|cc|cc|cc|cc|cc|cc|cc|}
      \hline
      \multicolumn{1}{|c|}{} 
        & \multicolumn{2}{c|}{\textbf{Presence (\%)}}
        & \multicolumn{4}{c|}{\textbf{Cell}}
        & \multicolumn{2}{c|}{\textbf{Row}}
        & \multicolumn{6}{c|}{\textbf{Table}} \\
      \cline{2-3} \cline{4-7} \cline{8-9} \cline{10-15}
      \textbf{Model}
        & \multicolumn{2}{c|}{}
        & \multicolumn{2}{c|}{F1}
        & \multicolumn{2}{c|}{Levenshtein}
        & \multicolumn{2}{c|}{F1}
        & \multicolumn{2}{c|}{Accuracy}
        & \multicolumn{2}{c|}{Levenshtein}
        & \multicolumn{2}{c|}{ROUGE-L} \\
      & U & S & U & S & U & S & U & S & U & S & U & S & U & S \\
      \hline
Qwen2.5-0.5B-Instruct & 99.66 & \textbf{100.00} & 0.724 & 0.519 & 0.804 & 0.753 & 0.231 & 0.060 & 0.217 & 0.060 & 0.888 & 0.864 & 0.871 & 0.831 \\
Qwen2.5-1.5B-Instruct & 99.34 & \textbf{100.00} & 0.845 & 0.763 & 0.912 & 0.860 & 0.436 & 0.295 & 0.436 & 0.295 & 0.943 & 0.908 & 0.943 & 0.898 \\
Qwen2.5-3B-Instruct & 99.91 & \textbf{100.00} & 0.822 & 0.832 & 0.876 & 0.906 & 0.383 & 0.431 & 0.381 & 0.431 & 0.939 & 0.930 & 0.936 & 0.937 \\
Qwen2.5-7B-Instruct & \underline{99.98} & \textbf{100.00} & \underline{0.888} & 0.787 & \underline{0.932} & 0.867 & \underline{0.561} & 0.314 & \underline{0.561} & 0.314 & 0.950 & 0.917 & 0.954 & 0.913 \\
Qwen2.5-14B-Instruct & \textbf{100.00} & \textbf{100.00} & \textbf{0.891} & 0.850 & \textbf{0.934} & 0.910 & \textbf{0.570} & 0.474 & \textbf{0.570} & 0.474 & \textbf{0.953} & 0.931 & \textbf{0.957} & 0.940 \\
Qwen2.5-32B-Instruct & \textbf{100.00} & 95.50 & 0.883 & 0.728 & 0.931 & 0.807 & 0.531 & 0.229 & 0.531 & 0.229 & \textbf{0.953} & 0.874 & 0.954 & 0.861 \\
Falcon3-1B-Instruct & 97.49 & \textbf{100.00} & 0.595 & 0.752 & 0.672 & 0.825 & 0.039 & 0.197 & 0.038 & 0.197 & 0.877 & 0.905 & 0.822 & 0.888 \\
Falcon3-3B-Instruct & 99.49 & \textbf{100.00} & 0.871 & 0.758 & 0.920 & 0.860 & 0.507 & 0.279 & 0.507 & 0.279 & 0.949 & 0.911 & 0.951 & 0.903 \\
Falcon3-7B-Instruct & 99.96 & \textbf{100.00} & 0.880 & 0.836 & \underline{0.932} & 0.905 & 0.548 & 0.443 & 0.548 & 0.443 & \underline{0.952} & 0.930 & 0.954 & 0.933 \\
Falcon3-10B-Instruct & \textbf{100.00} & \textbf{100.00} & 0.881 & 0.786 & 0.930 & 0.866 & 0.547 & 0.295 & 0.546 & 0.295 & 0.951 & 0.917 & \underline{0.955} & 0.910 \\
Phi-4-mini-Instruct & \textbf{100.00} & \textbf{100.00} & 0.838 & 0.707 & 0.885 & 0.846 & 0.439 & 0.208 & 0.439 & 0.208 & 0.944 & 0.900 & 0.944 & 0.890 \\
Phi-4 & \textbf{100.00} & \textbf{100.00} & 0.818 & 0.727 & 0.850 & 0.828 & 0.401 & 0.198 & 0.401 & 0.198 & 0.945 & 0.902 & 0.946 & 0.889 \\
      \hline
    \end{tabular}%
  }
    \caption{Evaluation metrics on the E2E test set. For each metric, Unstructured (U) and Structured (S) results are shown with the best model in bold and second best model underlined.}
        \label{tab:e2e-metrics}

\end{table*}

The Cell Level metrics, Levenshtein and F1, show in general higher values than the ones for row and table level. In comparison to Levenshtein and Rouge-L on Table Level we see - apart from the outlier - a clear expressed gain in performance with rising parameter sizes. 

\subsection{Rotowire}

\begin{table*}[h]
  \centering
  \small
  \renewcommand{\arraystretch}{1.2}
  \resizebox{\textwidth}{!}{%
    \begin{tabular}{
      |l|cc|cc|cc|cc|cc|cc|cc|cc|
    }
      \hline
      \multicolumn{1}{|c|}{} 
        & \multicolumn{2}{c|}{\textbf{Presence (\%)}}
        & \multicolumn{6}{c|}{\textbf{Cell}}
        & \multicolumn{2}{c|}{\textbf{Row}}
        & \multicolumn{6}{c|}{\textbf{Table}} \\
      \cline{2-3} \cline{4-9} \cline{10-11} \cline{12-17}
      \textbf{Model}
        & \multicolumn{2}{c|}{}
        & \multicolumn{2}{c|}{RMSE}
        & \multicolumn{2}{c|}{F1}
        & \multicolumn{2}{c|}{Levenshtein}
        & \multicolumn{2}{c|}{F1}
        & \multicolumn{2}{c|}{Exact Match}
        & \multicolumn{2}{c|}{Lev}
        & \multicolumn{2}{c|}{ROUGE-L} \\
      & U & S & U & S & U & S & U & S & U & S & U & S & U & S & U & S \\
      \hline
      Qwen2.5-0.5B-Instruct & 98.4 & 99.7 & 33.16 & 35.94 & 0.597 & 0.722 & 0.634 & 0.719 & 0.025 & 0.026 & 0.010 & 0.020 & 0.888 & 0.937 & 0.723 & 0.739 \\
      Qwen2.5-1.5B-Instruct & \textbf{100.0} & \textbf{100.0} & 39.57 & 17.77 & 0.619 & 0.859 & 0.581 & 0.847 & 0.042 & 0.140 & 0.031 & 0.090 & 0.949 & 0.961 & 0.874 & 0.839 \\
      Qwen2.5-3B-Instruct & \underline{99.9} & \textbf{100.0} & 43.85 & 14.35 & 0.595 & 0.883 & 0.549 & 0.853 & 0.042 & 0.194 & 0.031 & 0.108 & 0.962 & 0.964 & 0.917 & 0.857 \\
      Qwen2.5-7B-Instruct & 99.7 & \textbf{100.0} & 38.85 & 9.72 & 0.639 & 0.921 & 0.563 & 0.878 & 0.108 & 0.346 & 0.068 & 0.194 & 0.972 & 0.976 & 0.946 & 0.904 \\
      Qwen2.5-14B-Instruct & \textbf{100.0} & \textbf{100.0} & 33.68 & \underline{6.33} & 0.764 & \underline{0.964} & 0.705 & \textbf{0.939} & 0.105 & \underline{0.605} & 0.038 & \underline{0.426} & 0.983 &  \textbf{0.989} & \textbf{0.969} & 0.959 \\
      Qwen2.5-32B-Instruct & 96.9 & 99.4 & 41.40 & \textbf{6.05} & 0.623 & \textbf{0.965} & 0.540 & \textbf{0.939} & 0.056 & \textbf{0.616} & 0.017 & \textbf{0.433} & 0.964 & \textbf{0.989} & 0.952 & 0.958 \\
      Falcon3-1B-Instruct & 96.2 & 85.7 & 51.16 & 37.38 & 0.370 & 0.704 & 0.389 & 0.780 & 0.015 & 0.026 & 0.015 & 0.020 & 0.838 & 0.923 & 0.686 & 0.724 \\
      Falcon3-3B-Instruct & 98.8 & 96.5 & 48.25 & 30.98 & 0.501 & 0.777 & 0.431 & 0.824 & 0.021 & 0.066 & 0.019 & 0.048 & 0.943 & 0.946 & 0.885 & 0.776 \\
      Falcon3-7B-Instruct & \textbf{100.0} & \textbf{100.0} & 35.69 & 10.39 & 0.730 & 0.928 & 0.670 & 0.886 & 0.074 & 0.394 & 0.033 & 0.234 & 0.980 & 0.980 & 0.957 & 0.923 \\
      Falcon3-10B-Instruct & \textbf{100.0} & \textbf{100.0} & 47.07 & 7.52 & 0.595 & 0.951 & 0.501 & \underline{0.919} & 0.031 & 0.539 & 0.020 & 0.374 & 0.974 & \underline{0.986} & 0.966 & 0.948 \\
      Phi-4-mini-Instruct & \textbf{100.0} & 80.3 & 33.20 & 32.53 & 0.750 & 0.628 & 0.693 & 0.553 & 0.063 & 0.082 & 0.031 & 0.051 & 0.975 & 0.951 & 0.936 & 0.807 \\
      Phi-4 & \textbf{100.0} & \textbf{100.0} & 47.80 & 6.44 & 0.675 & 0.937 & 0.527 & 0.899 & 0.018 & 0.600 & 0.015 & 0.405 & 0.967 & \textbf{0.989} & \underline{0.968} & 0.957 \\
      \hline
    \end{tabular}%
  }
    \caption{Evaluation metrics on the Team Tables of the Rotowire test set. For each metric, Unstructured (U) and Structured (S) results are shown with the best model in bold and second best model underlined.}
  \label{tab:rotowire-team-metrics}
\end{table*}

\begin{table*}[h]
  \centering
  \small
  \renewcommand{\arraystretch}{1.2}
  \resizebox{\textwidth}{!}{%
    \begin{tabular}{
      |l|cc|cc|cc|cc|cc|cc|cc|cc|
    }
      \hline
      \multicolumn{1}{|c|}{} 
        & \multicolumn{2}{c|}{\textbf{Presence (\%)}}
        & \multicolumn{6}{c|}{\textbf{Cell}}
        & \multicolumn{2}{c|}{\textbf{Row}}
        & \multicolumn{6}{c|}{\textbf{Table}} \\
      \cline{2-3} \cline{4-9} \cline{10-11} \cline{12-17}
      \textbf{Model}
        & \multicolumn{2}{c|}{}
        & \multicolumn{2}{c|}{RMSE}
        & \multicolumn{2}{c|}{F1}
        & \multicolumn{2}{c|}{Levenshtein}
        & \multicolumn{2}{c|}{F1}
        & \multicolumn{2}{c|}{Accuracy}
        & \multicolumn{2}{c|}{Levenshtein}
        & \multicolumn{2}{c|}{ROUGE-L} \\
      & U  & S & U & S & U & S & U & S & U & S & U & S & U & S & U & S \\
      \hline
      Qwen2.5-0.5B-Instruct & 43.1 & \underline{99.4} & 11.47 & 10.98 & 0.661 & 0.697 & 0.550 & 0.633 & 0.001 & 0.022 & 0.000 & 0.006 & 0.684 & 0.925 & 0.522 & 0.674 \\
      Qwen2.5-1.5B-Instruct & 70.9 & \textbf{100.0} & 8.97 & 7.97 & 0.736 & 0.763 & 0.630 & 0.683 & 0.054 & 0.105 & 0.004 & 0.025 & 0.854 & 0.936 & 0.706 & 0.747 \\
      Qwen2.5-3B-Instruct & 88.5 & \textbf{100.0} & 6.96 & 4.97 & 0.751 & 0.898 & 0.655 & 0.835 & 0.114 & 0.270 & 0.026 & 0.046 & 0.909 & 0.945 & 0.769 & 0.798 \\
      Qwen2.5-7B-Instruct & 95.3 & \textbf{100.0} & 6.19 & 3.22 & 0.745 & 0.938 & 0.638 & 0.894 & 0.133 & 0.455 & 0.021 & 0.112 & 0.939 & 0.954 & 0.846 & 0.838 \\
      Qwen2.5-14B-Instruct & 96.3 & \textbf{100.0} & 4.99 & \underline{2.19} & 0.837 & \underline{0.964} & 0.756 & \underline{0.934} & 0.301 & \underline{0.623} & 0.061 & 0.188 & 0.962 & 0.955 & \underline{0.919} & 0.851 \\
      Qwen2.5-32B-Instruct & 77.1 &  \underline{99.4} & 5.12 & \textbf{1.78} & 0.794 & \textbf{0.971} & 0.689 & \textbf{0.946} & 0.223 & \textbf{0.687} & 0.037 & \textbf{0.224} & 0.892 & 0.957 & 0.843 & 0.859 \\
      Falcon3-1B-Instruct & 35.6 & 84.1 & 12.58 & 14.74 & 0.649 & 0.536 & 0.526 & 0.532 & 0.002 & 0.013 & 0.001 & 0.006 & 0.690 & 0.872 & 0.534 & 0.606 \\
      Falcon3-3B-Instruct & 92.7 & 96.7 & 6.70 & 7.15 & 0.663 & 0.656 & 0.569 & 0.585 & 0.102 & 0.059 & 0.015 & 0.021 & 0.914 & 0.933 & 0.775 & 0.699 \\
      Falcon3-7B-Instruct & 95.2 & \textbf{100.0} & 4.79 & 3.48 & 0.854 & 0.924 & 0.778 & 0.874 & 0.271 & 0.433 & 0.059 & 0.119 & 0.957 & 0.954 & 0.879 & 0.835 \\
      Falcon3-10B-Instruct & 94.2 & \textbf{100.0} & 5.88 & 2.41 & 0.780 & 0.961 & 0.673 & 0.930 & 0.186 & 0.609 & 0.026 & 0.189 & 0.940 & 0.956 & 0.909 & 0.851 \\
      Phi-4-mini-Instruct & 87.2 & 81.1 & 4.96 & 7.30 & 0.859 & 0.826 & 0.779 & 0.731 & 0.268 & 0.203 & 0.038 & 0.044 & \textbf{0.972} & 0.952 & 0.884 & 0.784 \\
      Phi-4 & \textbf{100.0} & \textbf{100.0} & 6.11 & 2.01 & 0.838 & 0.956 & 0.733 & 0.923 & 0.164 & 0.634 & 0.006 & \underline{0.196} & \underline{0.970} & 0.956 & \textbf{0.956} & 0.852 \\
      \hline
    \end{tabular}%
  }
    \caption{Evaluation metrics on the Player tables of the Rotowire dataset for Unstructured (U) vs. Structured (S) generation with the best model in bold and second best model underlined.}
  \label{tab:rotowire-player-metrics}
\end{table*}

The evaluation results (Table~\ref{tab:rotowire-team-metrics}, Table~\ref{tab:rotowire-player-metrics}) demonstrate that guided (structured) decoding consistently improves model performance on the Rotowire dataset. Across both Team and Player tables, all evaluated models achieve high table presence rates, with structured decoding frequently reaching or approaching 100\%. Notably, the smallest model (Qwen2.5-0.5B-Instruct) generates only 43\% of player tables in the unstructured setting, but rises to 99.4\% with guided decoding. For all core metrics - RMSE, cell F1, cell cell Levenshtein, row F1, table exact match, Levenshtein, and ROUGE-L - structured outputs generally yield equal or higher scores than unstructured ones. The lowest RMSEs are observed for the largest Qwen2.5 models with structured decoding (e.g., 1.78 for Player, 6.05 for Team). Both cell F1 and Lev-F1 scores are maximized in structured outputs of larger Qwen2.5 and Falcon models, frequently exceeding 0.96.

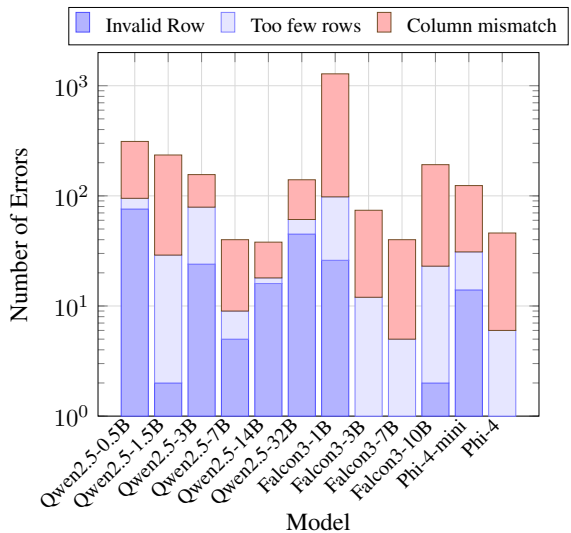
\begin{figure}[h!]
  \centering
  \resizebox{\columnwidth}{!}{
    \begin{tikzpicture}
  \begin{axis}[
      ymode=log,
      ymin=1,
      ymax=2000,
      ylabel={Number of Errors},
      xlabel={Model},
      xlabel style={at={(axis description cs:0.5,-0.15)},anchor=north},
      symbolic x coords={
        Qwen2.5-0.5B,Qwen2.5-1.5B,Qwen2.5-3B,Qwen2.5-7B,
        Qwen2.5-14B,Qwen2.5-32B,Falcon3-1B,Falcon3-3B,
        Falcon3-7B,Falcon3-10B,{Phi-4-mini},{Phi-4}
      },
      xtick=data,
      x tick label style={rotate=45,anchor=east,font=\small},
      bar width=12pt,
           legend style={
            at={(0.5,1.02)},
            anchor=south,
            legend columns=3,
            font=\footnotesize,
            column sep=1ex
          },
      grid=major,
      grid style={gray!30}
    ]

    \addplot+[ybar stacked, mark=none, draw=blue!70,  fill=blue!30] coordinates {
      (Qwen2.5-0.5B,   76)  (Qwen2.5-1.5B,   2)   (Qwen2.5-3B,  24)
      (Qwen2.5-7B,      5)  (Qwen2.5-14B,   16)  (Qwen2.5-32B, 45)
      (Falcon3-1B,  26)  (Falcon3-3B,  1)   (Falcon3-7B, 1)
      (Falcon3-10B,  2)  (Phi-4-mini, 14)   (Phi-4,     1)
    };
    \addlegendentry{Invalid Row}

    \addplot+[ybar stacked, mark=none, draw=blue!50,  fill=blue!10] coordinates {
      (Qwen2.5-0.5B,   19)  (Qwen2.5-1.5B,  27)  (Qwen2.5-3B,  55)
      (Qwen2.5-7B,      4)  (Qwen2.5-14B,    2)  (Qwen2.5-32B, 16)
      (Falcon3-1B,  72)  (Falcon3-3B, 11)  (Falcon3-7B, 4)
      (Falcon3-10B,21)  (Phi-4-mini, 17)  (Phi-4,     5)
    };
    \addlegendentry{Too few rows}

    \addplot+[ybar stacked, mark=none draw=red!50,   fill=red!30] coordinates {
      (Qwen2.5-0.5B,   217) (Qwen2.5-1.5B, 206) (Qwen2.5-3B,   77)
      (Qwen2.5-7B,      31) (Qwen2.5-14B,   20) (Qwen2.5-32B,  79)
      (Falcon3-1B,1184) (Falcon3-3B,  62) (Falcon3-7B, 35)
      (Falcon3-10B,169) (Phi-4-mini, 93) (Phi-4,     40)
    };
    \addlegendentry{Column mismatch}

  \end{axis}
\end{tikzpicture}
  }
  \caption{Distribution of errors for parsing the table candidates from unstructured generation on the Rotowire dataset. `Invalid row' errors combine all errors with invalid separator line, invalid headers and invalid data rows. Tables are classified as `Too few rows' with less than 3 rows and `Column mismatch' occurs when the amount of columns over the table rows does not align.}
  \label{fig:error-stacks}
\end{figure}

An analysis of the errors that occured, when validating the table candidates in the unstructured outputs show (Figure \ref{fig:error-stacks}), that by far the most common error type is the column missmatch. For all different model families these error is reduced constantly with an increase of model size, however the Qwen and the Falcon family show a rise of candidate errors for the biggest evaluated models (32B and 10B respectively). When comparing Falcon3-7B and Falcon3-10B, it is to note, that while the table presences only drop by 1\%, the amount of candidate errors rises overproportional.

\begin{figure*}[htbp]
\centering
\small
\begin{subfigure}[b]{0.35\textwidth}
    \centering
    \begin{tikzpicture}
\begin{axis}[
    width=\textwidth,
    height=6cm,
    ybar,
    bar width=7pt,
    ymin=0,
    ymax=55,
    ylabel={RMSE},
    ylabel style={yshift=-5ex},
    xlabel={Parameter Size},
    symbolic x coords={0.5B,1.5B,3B,7B,14B,32B},
    xtick=data,
    legend style={at={(0.5,-0.2)}, anchor=north,legend columns=2},
    nodes near coords,
    every node near coord/.append style={font=\scriptsize, rotate=90, anchor=west, yshift=0.2ex},
    enlarge x limits=0.1,   
    grid=major,
    legend image code/.code={
      \draw[draw=#1,line width=1pt] (0cm,-0.1cm) rectangle (0.4cm,0.2cm);
    }
]
\addplot+[draw=blue, fill=blue!30, bar shift=-3.5pt, nodes near coords, every node near coord/.append style={blue}] coordinates {
    (0.5B,33.16) (1.5B,39.57) (3B,43.85) (7B,38.85) (14B,33.68) (32B,41.40)
};
\addplot+[draw=red, fill=gray!40, bar shift=+3.5pt, nodes near coords, every node near coord/.append style={red}] coordinates {
    (0.5B,35.94) (1.5B,17.77) (3B,14.35) (7B,9.72) (14B,6.33) (32B,6.05)
};
\end{axis}
\end{tikzpicture}
    \caption{Rotowire Team}
\end{subfigure}\hspace{-1.5em}
\begin{subfigure}[b]{0.35\textwidth}
    \centering
    \begin{tikzpicture}
\begin{axis}[
    width=\textwidth,
    height=6cm,
    ybar,
    bar width=7pt,
    ymin=0,
    ymax=15,
    ylabel={RMSE},
    ylabel style={yshift=-5ex},
    xlabel={Parameter Size},
    symbolic x coords={0.5B,1.5B,3B,7B,14B,32B},
    xtick=data,
    legend style={at={(0.5,1.07)}, anchor=south, legend columns=2},
    nodes near coords,
    every node near coord/.append style={font=\scriptsize, rotate=90, anchor=west, yshift=0.7ex},
    enlarge x limits=0.1,
    grid=major,
    legend image code/.code={
      \draw[draw=#1,line width=1pt] (0cm,-0.1cm) rectangle (0.4cm,0.2cm);
    }
]
\addplot+[draw=blue, fill=blue!30, bar shift=-3.5pt, nodes near coords, every node near coord/.append style={blue}] coordinates {
    (0.5B,11.47) (1.5B,8.97) (3B,6.96) (7B,6.19) (14B,4.99) (32B,5.12)
};
\addlegendentry{Unstructured}
\addplot+[draw=red, fill=gray!40, bar shift=+3.5pt, nodes near coords, every node near coord/.append style={red}] coordinates {
    (0.5B,10.98) (1.5B,7.97) (3B,4.97) (7B,3.22) (14B,2.19) (32B,1.78)
};
\addlegendentry{Structured}
\end{axis}
\end{tikzpicture}
    \caption{Rotowire Player}
\end{subfigure}\hspace{-1.5em}
\begin{subfigure}[b]{0.35\textwidth}
    \centering
    \begin{tikzpicture}
\begin{axis}[
    width=\textwidth,
    height=6cm,
    ybar,
    bar width=7pt,
    ymin=0,
    ymax=9,
    ylabel={RMSE},
    ylabel style={yshift=-5ex},
    xlabel={Parameter Size},
    symbolic x coords={0.5B,1.5B,3B,7B,14B,32B},
    xtick=data,
    nodes near coords,
    every node near coord/.append style={font=\scriptsize, rotate=90, anchor=west, yshift=0.2ex},
    enlarge x limits=0.1,
    grid=major,
    legend image code/.code={
      \draw[draw=#1,line width=1pt] (0cm,-0.1cm) rectangle (0.4cm,0.2cm);
    }
]
\addplot+[draw=blue, fill=blue!30, bar shift=-3.5pt, nodes near coords, every node near coord/.append style={blue}] coordinates {
    (0.5B,3.49) (1.5B,3.66) (3B,3.11) (7B,2.91) (14B,2.32) (32B,2.17)
};
\addplot+[draw=red, fill=gray!40, bar shift=+3.5pt, nodes near coords, every node near coord/.append style={red}] coordinates {
    (0.5B,7.64) (1.5B,6.37) (3B,2.84) (7B,3.08) (14B,2.81) (32B,2.83)
};
\end{axis}
\end{tikzpicture}
    \caption{Livesum}
\end{subfigure}
\caption{The influence of Qwen parameter size on the RMSE for unstructured vs. structured generation on the different table types with numerical cells.}
\label{fig:rmse}
\end{figure*}
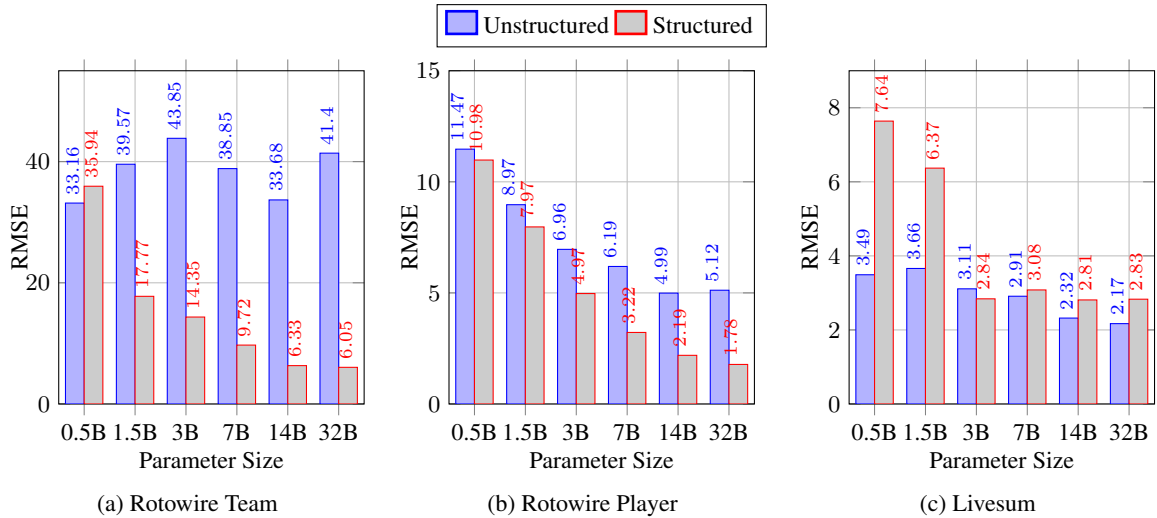

\subsection{Livesum}

Results on the Livesum dataset (Table~\ref{tab:livesum-team-metrics}) indicate that, while structured decoding increases the presence rate always to 100\%, it does not consistently improve table quality. For most cell-level metrics, unstructured outputs usually achieve better values. Importantly, none of the generated tables - regardless of decoding strategy - perfectly matches the ground truth, as evidenced by the absence of exact matches at both row and table levels.

\begin{table*}[h]
  \centering
  \small
  \renewcommand{\arraystretch}{1.2}
  \resizebox{\textwidth}{!}{%
    \begin{tabular}{
      |l|cc|cc|cc|cc|cc|cc|cc|cc|
    }
      \hline
      \multicolumn{1}{|c|}{} 
        & \multicolumn{2}{c|}{\textbf{Presence (\%)}}
        & \multicolumn{6}{c|}{\textbf{Cell}}
        & \multicolumn{2}{c|}{\textbf{Row}}
        & \multicolumn{6}{c|}{\textbf{Table}} \\
      \cline{2-3} \cline{4-9} \cline{10-11} \cline{12-17}
      \textbf{Model}
        & \multicolumn{2}{c|}{}
        & \multicolumn{2}{c|}{RMSE}
        & \multicolumn{2}{c|}{F1}
        & \multicolumn{2}{c|}{Levenshtein}
        & \multicolumn{2}{c|}{F1}
        & \multicolumn{2}{c|}{Accuracy}
        & \multicolumn{2}{c|}{Levenshtein}
        & \multicolumn{2}{c|}{ROUGE-L} \\
      & U & S & U & S & U & S & U & S & U & S & U & S & U & S & U & S \\
      \hline
Qwen2.5-0.5B-Instruct & 99.7 & \textbf{100.0} & 3.49 & 7.64 & 0.481 & 0.460 & 0.768 & 0.755 & 0.000 & 0.000 & 0.000 & 0.000 & 0.925 & 0.922 & 0.674 & 0.625 \\
Qwen2.5-1.5B-Instruct & 99.6 & \textbf{100.0} & 3.66 & 6.37 & 0.549 & 0.403 & 0.789 & 0.516 & 0.000 & 0.000 & 0.000 & 0.000 & 0.931 & 0.936 & 0.693 & 0.690 \\
Qwen2.5-3B-Instruct & 96.7 & \textbf{100.0} & 3.11 & 2.84 & 0.562 & 0.586 & 0.798 & 0.783 & 0.000 & 0.000 & 0.000 & 0.000 & 0.932 & 0.937 & 0.695 & 0.714 \\
Qwen2.5-7B-Instruct & 98.5 & \textbf{100.0} & 2.91 & 3.08 & 0.490 & 0.571 & 0.777 & 0.773 & 0.000 & 0.000 & 0.000 & 0.000 & 0.928 & 0.935 & 0.683 & 0.710 \\
Qwen2.5-14B-Instruct & 97.1 & \textbf{100.0} & \underline{2.32} & 2.81 & 0.647 & 0.581 & \underline{0.826} & 0.727 & 0.000 & 0.000 & 0.000 & 0.000 & \underline{0.942} & 0.938 & 0.735 & 0.727 \\
Qwen2.5-32B-Instruct & \underline{99.9} & \textbf{100.0} & \textbf{2.17} & 2.83 & \textbf{0.670} & 0.573 & \textbf{0.837} & 0.693 & \textbf{0.002} & 0.000 & 0.000 & 0.000 & \textbf{0.944} & 0.939 & \textbf{0.747} & 0.728 \\
Falcon3-1B-Instruct & 82.1 & \textbf{100.0} & 3.99 & 7.56 & 0.440 & 0.308 & 0.727 & 0.713 & 0.000 & 0.000 & 0.000 & 0.000 & 0.900 & 0.879 & 0.653 & 0.574 \\
Falcon3-3B-Instruct & 92.3 & \textbf{100.0} & 3.42 & 5.41 & 0.528 & 0.557 & 0.783 & 0.790 & 0.000 & 0.000 & 0.000 & 0.000 & 0.929 & 0.932 & 0.690 & 0.706 \\
Falcon3-7B-Instruct & \textbf{100.0} & \textbf{100.0} & 3.13 & 5.34 & 0.577 & 0.439 & 0.803 & 0.515 & 0.000 & 0.000 & 0.000 & 0.000 & 0.934 & 0.924 & 0.701 & 0.693 \\
Falcon3-10B-Instruct & \textbf{100.0} & \textbf{100.0} & 2.79 & 4.27 & 0.579 & 0.587 & 0.803 & 0.750 & 0.000 & 0.000 & 0.000 & 0.000 & 0.934 & 0.936 & 0.702 & 0.728 \\
Phi-4-mini-Instruct & \textbf{100.0} & \textbf{100.0} & 3.47 & 6.86 & 0.595 & 0.472 & 0.796 & 0.505 & 0.000 & 0.000 & 0.000 & 0.000 & 0.933 & 0.939 & 0.718 & 0.717 \\
Phi-4 & \textbf{100.0} & \textbf{100.0} & 2.47 & 3.01 & \underline{0.648} & 0.555 & \underline{0.826} & 0.708 & 0.000 & 0.000 & 0.000 & 0.000 & \underline{0.942} & 0.934 & \underline{0.736} & 0.720 \\
      \hline
    \end{tabular}%
  }
  \caption{Evaluation metrics on the Livesum test set. For each metric, Unstructured (U) and Structured (S) results are shown with the best model in bold and second best model underlined.}
    \label{tab:livesum-team-metrics}

\end{table*}


\section{Discussion}

\subsection{Decoding Strategy across Tasks}

Our results proved, that structured decoding consistently boosts the \emph{presence} of valid tables across all three benchmarks: malformed outputs are far less common than with one-shot prompting.  Beyond mere presence, it improves table \emph{quality} on both Rotowire tasks (Team and Player), whereas it is counterproductive on E2E and Livesum. On the Rotowire benchmark, the schema eliminates most errors: even the smallest model (Qwen2.5-0.5B) shows an increased performance once structured decoding is applied, and the gains grow significantly with parameter size. In contrast, unstructured generation struggles on team tables to reliably align values with the correct entities, likely due to the dense and entangled presentations of statistics for both teams in the source text. The picture flips on the E2E dataset: critical attributes are densely packed within short utterances, and the freedom of unconstrained decoding lets larger models capture subtle lexical cues better than a rigid schema.
The results for Livesum show, that even though structured decoding guarantees full coverage of the tables it does also not raise the quality metrics. Here we assume that the reason is due to the high reasoning requiring task and aggregating information over long contexts, a task where already \citet{tam-etal-2024-speak} showed, that reasoning skills are lowered, when enforcing high structural constrained. 

In short, strict schema-guided decoding is helpful when numerical information can be directly extracted from sparsely spread information in the text (Rotowire) but can hinder performance when textual information is densely packed (E2E). For Livesum, the models must infer final values by aggregating evidence scattered across long articles, a reasoning-heavy task that neither decoding strategy handles well.

\subsection{Influence of Model Size}

Overall, we observe a clear positive relationship between model size and the validity and quality of generated tables. However, this trend does not hold uniformly across all datasets and decoding methods. Notably, our largest evaluated model, Qwen2.5-32B, demonstrates superior performance on the Livesum and Rotowire datasets according to most metrics, yet it unexpectedly shows reduced table presence, particularly pronounced on the Rotowire dataset in both decoding settings, and on the E2E dataset with structured decoding. It also showed significantly reduced table quality with respect to the metrics, than smaller models of the same family on E2E. For tables containing numerical values, we further investigated the RMSE to better reflect true table fidelity, as string-based metrics may provide misleading interpretations here. Larger models typically yielded lower RMSE values, however, the Rotowire Team Tables showed consistently high RMSE under unstructured decoding, regardless of model size, which suggests limitations specific to that scenario. In contrast, structured decoding consistently improved RMSE performance on these tables.

In summary, while larger models generally enhance table generation quality and validity, our work also showed exceptions, particularly Qwen2.5-32B, where increased model size adversely affects table presence and certain performance aspects. 

\subsection{Suitability of Evaluation Metrics}

Our findings indicate a limited suitability of common NLP metrics at the table level. Despite high Levenshtein ratios and ROUGE-L scores, the models never achieve exact row- or table-level matches, indicating that such metrics may overestimate the true quality of the generated tables. A similar observation can be made for finegrained cell level metrics, such as the F1 Score or Cell-level Levenshtein. In contrast, RMSE provides a informative assessment for numerical tables, directly quantifying the deviation from ground truth, but is not applicable to tables with textual cells. The strict exact match metrics at row and table level accurately indicate whether the generated table matches the ground truth, but they fail to account for semantically equivalent variations or minor deviations in string expression. The Levenshtein score on positional cell level has been found useful, as it clearer expresses the differences in performance between the different model sizes, while not being as strict as the exact match metrics. In general the metrics were able to consistently indicate, whether structured decoding performed better or worse on a given benchmark; however, due to their individual limitations, actual table quality is often best assessed through human evaluation. 

\subsection{Limitations}
All our benchmarks assume known schemas for table generation; we do not address open-schema or schema-inference scenarios, as explored in recent work~\cite{ahuja2025mapmakeschemaguidedtext}. Our experiments are also limited to a specific one-shot prompt and a structured decoding prompt, and do not consider alternative prompting strategies that might yield better performance. Additionally, we evaluate only a subset of available models, ranging from 0.5B to 32B parameters, from three developers (tiiuae, Qwen, Microsoft), none of whom publish their training data. As a result, we cannot rule out the possibility that some models may have been exposed to our benchmark datasets during training. Furthermore the selected benchmarks represent just a limited range of domains and table types. Our results may not generalize to other datasets, especially those with more complex tables, larger sizes, or more diverse content. Our evaluation also assumes accurate table extraction and preprocessing from ground truth and LLM responses. Any errors or inconsistencies in preprocessing could impact the reported metrics.

 \section{Conclusion \& Future Work}

 Our study demonstrated that the impact of structured decoding on information extraction in the context of Text-to-Table generations is \emph{highly task dependent}. We derived the following key conclusions:
 \begin{itemize}
     \item \textbf{Decoding strategy:} Schema-guided decoding is improving the presence of tables and reducing malformed outputs significantly, but it depresses the table quality where textual facts are densely packed within the input or have to be aggregated over long context.
     \item \textbf{Model size:} The relationship between model size and table generation quality follows predictable trends in most cases, with larger models generally producing higher validity and quality tables. Results on the Rotowire Team tables however show for the unstructured setting, that scale alone does not guarantee optimal performance for table generation tasks.
     \item \textbf{Evaluation Metrics:} String-based NLP metrics on Table level overestimate table quality. While exact match metrics are to strict for probabilistic generated content, they reflect the actual table qualities better. A mixed variant, utilizing NLP based soft-match metrics positionally on cell or row level, seems more promising. 
 \end{itemize}
 
Considering the limitations identified in our work, we recommend several promising directions for future research. First, there is significant value in exploring methods for schema inference, moving beyond the exclusive use of predefined schemas. Further research should also address the generation of more complex tables, such as those exhibiting greater variability in size, multi-line headers, or merged cells. Additionally, we encourage the development of advanced methods for constrained decoding that go beyond the application of standard JSON schemas. For instance, XML-grammars or the design and implementation of table-specific grammars tailored to the target table sequence language, alongside systematic evaluation of their computational efficiency. Finally, we emphasize the need for novel evaluation metrics and Text-to-Table datasets, ideally complemented by human assessments, to more robustly measure the effectiveness of generated tables and better capture the nuances of real-world use cases.

\section*{Acknowledgments}
This work has been partially funded by the German Federal Ministry of Research, Technology, and Space (BMFTR) under the grant numbers 01IS24037B and 01IS24077A. Computations for this work were done (in part) using resources of the Leipzig University Computing Centre.

\bibliographystyle{acl_natbib}
\bibliography{ranlp2025}

\end{document}